\documentclass[runningheads]{llncs}
\usepackage[T1]{fontenc}
\usepackage{graphicx}
\usepackage[table]{xcolor}
\usepackage{colortbl} 
\usepackage{multirow}
\usepackage{tabularx}
\usepackage{booktabs}
\usepackage{geometry} 
\usepackage{pdflscape}
\usepackage[hyphens]{url}
\usepackage{xurl}

\usepackage[
  colorlinks=true,
  linkcolor=blue,
  urlcolor=blue,
  breaklinks=true   
]{hyperref}

\usepackage{pgfplots}
\usepgfplotslibrary{groupplots,statistics}
\pgfplotsset{compat=1.17}

\usepackage{enumitem}
\setlist[itemize]{label={\textperiodcentered}}
\usepackage{orcidlink}

\newcommand{\orcidauthorE}{0000-0002-3658-6782}

\usepackage[normalem]{ulem} 

\begin{document}
\title{Distinguishing AI-Generated and Human-Written Text Through Psycholinguistic Analysis}
%

\author{Chidimma Opara\orcidlink{\orcidauthorE}}
\authorrunning{C. Opara}

\institute{School of Computing, Engineering \& Digital Technologies\\
Teesside University, Middlesbrough, UK\\
\email{c.opara@tees.ac.uk}}

\maketitle              
\begin{abstract}
The increasing sophistication of AI-generated texts highlights the urgent need for accurate and transparent detection tools, especially in educational settings, where verifying authorship is essential. Existing literature has demonstrated that the application of stylometric features with machine learning classifiers can yield excellent results. Building on this foundation, this study proposes a comprehensive framework that integrates stylometric analysis with psycholinguistic theories, offering a clear and interpretable approach to distinguishing between AI-generated and human-written texts.
This research specifically maps 31 distinct stylometric features to cognitive processes such as lexical retrieval, discourse planning, cognitive load management, and metacognitive self-monitoring. In doing so, it highlights the unique psycholinguistic patterns found in human writing. Through the intersection of computational linguistics and cognitive science, this framework contributes to the development of reliable tools aimed at preserving academic integrity in the era of generative AI.

\end{abstract}

\section{Introduction}
\label{sec:introduction}

AI-generated texts, powered by advanced models such as Openai’s GPT series, have significantly reshaped content creation, enabling artificial intelligence to produce text that closely resembles human writing. While these systems enhance efficiency and creativity, such as improving learning experiences \cite{ngo2023perception}, they also introduce new challenges in distinguishing AI-generated content from human-authored work, particularly in education, where written assignments serve as a key measure of critical thinking and subject expertise \cite{bailey2009undergraduate}.
In response to these challenges, recent state-of-the-art research has focused on AI-text detection using deep learning techniques, including Convolutional Neural Networks, Recurrent Neural Networks, Long Short-Term Memory networks, and Transformer-based models such as RoBERTa and GPT. These models have demonstrated excellent performance in classification tasks \cite{sarvazyan2023overview}. However, they are often criticised for their “black box” nature, offering high accuracy without transparent reasoning.

A more interpretable and linguistically grounded approach is \textbf{stylometry}, the quantitative analysis of writing style, traditionally used in literary authorship attribution. Stylometric techniques analyse measurable features such as lexical diversity, sentence complexity, and readability to identify patterns indicative of authorship \cite{opara2024styloai,kumarage2023stylometric,mindner2023classification}. Although stylometric features often yield high classification accuracy when applied on machine learning algorithms, they typically lack explanatory capacity regarding the cognitive processes that give rise to these features in human writing.

Understanding human authorship requires a deeper engagement with the \textbf{psycholinguistic} mechanisms that shape written expression. Stylometric features in human-authored texts are not arbitrary; they reflect complex cognitive activities such as self-monitoring, discourse planning, and lexical retrieval \cite{garcia2024evaluating,huang2025authorship}. For instance, higher syntactic complexity may indicate intentional structuring of ideas. By aligning stylometric analysis with psycholinguistic theories, it becomes possible to develop more interpretable and cognitively informed approaches to distinguishing human-authored and AI-generated text.

\subsection{Research Contribution}
This study builds upon prior research \cite{opara2024styloai} and extends the AI-generated text detection model (\href{https://scholar.google.co.uk/citations?view_op=view_citation&hl=en&user=CXPDefcAAAAJ&citation_for_view=CXPDefcAAAAJ:Y0pCki6q_DkC}{StyloAI}) by grounding stylometric features in psycholinguistic theory. 31 stylometric features are systematically mapped to cognitive and linguistic processes, including Cognitive Load Theory, Metacognition and Self-Monitoring, Lexical Access and Retrieval, and Discourse Planning and Cohesion. This theoretical alignment enhances the understanding of why each of the stylometric features serves as an effective discriminator between human and AI-generated writing.

The remainder of the paper is organised as follows: The next section reviews psycholinguistic theories relevant to authorship detection. Section \ref{sys-analysis} provides an in-depth analysis of six stylometric feature categories: Lexical, Syntactic, Sentiment, Readability, Named Entity, and Uniqueness, each linked to corresponding psycholinguistic processes. Section \ref{sec:conclusion} concludes the paper.

\section{Psycholinguistic Theories in Human and AI-Generated Writing} \label{sec:background}
\subsection{\textbf{Cognitive Load Theory}}
Cognitive Load Theory (CLT), introduced by Sweller in 1988 \cite{sweller1994cognitive}, posits that human working memory has limited capacity. When writing tasks overwhelm this capacity, through the simultaneous demands of idea generation, sentence construction, and argument organisation, performance declines and error likelihood increases \cite{baddeley2003working}.
This strain manifests in natural markers such as pauses, revisions, and stylistic fluctuations, which contribute to the uniqueness of human writing. For instance, under increased cognitive load, writers may simplify sentence structures or omit details \cite{goldstein2022shared}. These features are detectable through stylometric analysis and offer insight into the cognitive processes involved in human authorship.

By contrast, AI systems do not experience cognitive load in the human sense. Large Language Models (LLMs) generate coherent, fluent text without pausing, self-monitoring, or strategic planning \cite{suresh2023conceptual}. While AI is bound by computational limits, these do not mirror the constraints of human cognition. As a result, AI-generated writing tends to be polished and consistent, yet often lacks the subtle imperfections typical of human expression. Efforts to simulate human-like limitations, such as bounded pragmatic speaker models \cite{nguyen2023language}, remain artificial and do not replicate authentic cognitive processes.

\subsection{\textbf{Metacognition and Self-Monitoring}}
Metacognition, as defined by Flavell (1979) \cite{flavell1979theories}, refers to one’s ability to reflect on and regulate cognitive processes. In writing, this includes revising content to improve clarity, coherence, and alignment with audience expectations. Such self-monitoring is evident in abrupt stylistic changes, iterative corrections, and context-sensitive modifications during drafting.

AI models, by contrast, lack fundamental metacognitive capabilities. Although models like GPT-4 can simulate reasoning through techniques such as chain-of-thought prompting \cite{collins2022structured,mahowald2024dissociating}, they do not engage in conscious reflection or deliberate revision. Instead, their refinements are statistical, not intentional. This limitation is especially evident in phenomena like hallucinations, overconfident but inaccurate outputs \cite{smith2023hallucination}. Unlike AI, human authors revise to correct meaning, adjust tone, or refine structure, making self-monitoring a key psycholinguistic marker of human authorship \cite{talboy2023challenging}.

\subsection{\textbf{Lexical Access and Retrieval}} Word selection in human writing involves a complex interplay between semantics, grammar, memory, and personal experience \cite{dell1992stages}. This cognitive process can result in occasional retrieval errors, such as tip-of-the-tongue moments or unintended repetition, often caught and corrected through self-monitoring. These subtle markers reflect real-time mental effort.

AI models, however, select words based on probabilistic associations derived from training data. While capable of producing text that mimics emotional richness, AI lacks lived experience and contextual depth, limiting the authenticity of its lexical choices \cite{mahowald2024dissociating}. Neuroimaging studies confirm that human writing activates brain areas related to memory and emotion dimensions AI cannot replicate \cite{goldstein2022shared,tuckute2024language}. Though AI may demonstrate lexical diversity, it often struggles to maintain tonal consistency and semantic cohesion over longer passages \cite{collins2022structured}, making lexical variation another useful marker for authorship attribution.

\subsection{\textbf{Discourse Planning and Cohesion}} Effective writing requires logical progression of ideas, as discussed by Halliday and Hasan (1976) \cite{carrell1982cohesion}. Human writers use cohesive devices such as pronouns, transitions, and structured argumentation to build narratives hierarchically: from broad ideas to fine-grained sentence-level refinements. Yet, under cognitive load, this planning may falter, leading to abrupt topic changes or weak transitions.

AI-generated texts emulate coherence via statistical predictions rather than conceptual planning \cite{mahowald2024dissociating,tuckute2024language}. While initially appearing structured, these texts often contain contradictions or repetitiveness due to shallow contextual modelling \cite{nolfi2024unexpected}. In contrast, human writing typically reflects more intentional and consistent discourse planning, making cohesion and logical progression key differentiators \cite{collins2022structured}.

\section{Framework Mapping Stylometric Features to Psycholinguistic Theories} \label{sys-analysis}

This section presents a systematic analysis of writing by integrating stylometric features with psycholinguistic theory. 

The \href{https://scholar.google.co.uk/citations?view_op=view_citation&hl=en&user=CXPDefcAAAAJ&citation_for_view=CXPDefcAAAAJ:Y0pCki6q_DkC}{StyloAI} model proposed in the previous work \cite{opara2024styloai} incorporates 31 stylometric features, including 12 novel metrics specifically developed for detecting AI-generated texts. These features are organised into six distinct categories: Lexical Diversity, Syntactic Complexity, Sentiment and Subjectivity, Readability, Named Entities, and Uniqueness and Variety. For an extensive discussion on the theoretical rationale and detailed feature descriptions, readers are referred to the \href{https://arxiv.org/html/2405.10129v1#:~:text=Table%201%3A,Across%20Six%20Categorie}{previous work} and Table \ref{tab:merged-stylometric-features} in the Appendix.

Table \ref{tab:colourful-features} summarises the mapping of 18 out of the 31 of these features to psycholinguistic theories.

\begin{table}[!ht]
\centering
\rowcolors{2}{blue!15}{white}
\begin{tabular}{|l|l|r|p{7cm}|}
\hline
\rowcolor{blue!50}
\textbf{Stylometric Category} & \textbf{Psycholinguistic Category} & \textbf{Count} & \textbf{Features} \\
\hline
Lexical Features             & Cognitive Load                         & 1 & word\_count \\
                              & Metacognition \& Self-Monitoring       & 1 & hapax\_legomenon\_rate \\
                              & Lexical Access \& Retrieval            & 4 & unique\_word\_count, ttr, avg\_word\_length, hapax\_legomenon\_rate \\
                              & Discourse Planning \& Cohesion         & 1 & char\_count \\
\hline
Syntactic Features           & Cognitive Load                         & 2 & avg\_sentence\_length, complex\_sentence\_count \\
                              & Metacognition \& Self-Monitoring       & 3 & punctuation\_count, question\_count, contraction\_count \\
                              & Discourse Planning \& Cohesion         & 1 & complex\_sentence\_count \\
\hline
Uniqueness Features          & Discourse Planning \& Cohesion         & 1 & syntax\_variety \\
\hline
Sentiment Features           & Cognitive Load                         & 1 & subjectivity \\
                              & Metacognition \& Self-Monitoring       & 2 & emotion\_word\_count, vader\_compound \\
\hline
Readability Features         & Cognitive Load                         & 1 & gunning\_fog \\
                              & Metacognition \& Self-Monitoring       & 1 & gunning\_fog \\
                              & Discourse Planning \& Cohesion         & 1 & flesch\_reading\_ease \\
\hline
Named Entity Features        & Metacognition \& Self-Monitoring       & 1 & first\_person\_count \\
                              & Discourse Planning \& Cohesion         & 1 & direct\_address\_count \\
\hline
\end{tabular}
\caption{ Mapping Stylometric Features to Psycholinguistic Theories.}
\label{tab:colourful-features}
\end{table}

\subsection{Detailed Feature Mapping}

\subsubsection{Lexical Features} encompass the fundamental elements of language, words, and their usage patterns. This category includes metrics such as total word count, vocabulary diversity, average word length, and the frequency of unique words. In academic contexts, these features provide insight into cognitive processes such as lexical retrieval, semantic depth, and the ability to structure and integrate knowledge effectively \cite{mindner2023classification,paas2020cognitive}.

\noindent  For instance, the following is an excerpt of a human-written text from the dataset used in the \href{https://arxiv.org/html/2405.10129v1#:~:text=Table%201%3A,Across%20Six%20Categorie}{StyloAI} model, that statistically has a high value of unique word count:
\begin{quote}
\texttt{"as you suspect a digestive problem in your rabbit, take him to your veterinarian immediately. If your rabbit has diarrhea, your veterinarian will test the feces to identify the specific organism (e.g., Clostridium)"}
\end{quote}

\paragraph{Psycholinguistic Interpretation:}
Human authors draw on semantic memory, lived experiences, and an awareness of audience expectations when choosing words. This often results in greater vocabulary diversity, precise language use, and contextually refined phrasing. Using stylometry, these behaviours are captured in features such as \textit{Unique Word Count}, \textit{TTR }and \textit{Hapax Legomenon rate}. Psycholinguistically, these behaviours reflect active \textit{lexical retrieval and metacognitive self-monitoring.}
In contrast, AI models, although capable of statistically simulating lexical variation, often struggle to produce contextualised or original word choices, especially across longer passages.

\subsubsection{Syntactic Features:}
Syntactic features examine sentence structure and grammatical complexity. Sentence construction places demands on working memory and discourse organisation, making it a key area where human and AI writing differ \cite{sweller1994cognitive}.

\paragraph{Psycholinguistic Interpretation:}
Human writers demonstrate flexibility in managing syntactic structure, tailoring sentence length, punctuation, and phrasing based on communicative goals such as clarity, emphasis, or engagement. These adjustments are enabled by all psycholinguistic theories discussed in Section \ref{sec:background}, especially \textit{self-monitoring} and \textit{discourse planning} processes that respond dynamically to context and audience.
In contrast, while syntactically fluent, AI-generated texts often exhibit a more rigid, repetitive structure. Without intrinsic self-monitoring, AI cannot vary syntax for rhetorical or communicative effect. As a result, its writing tends to lack the stylistic modulation typical of human authorship.

\subsubsection{Sentiment Features:}
Sentiment analysis evaluates the emotional tone conveyed in a text. For example, the \textit{Emotion Word Count}, which counts the total number of words associated with emotions in the text, is one of the stylometric features extracted in this category. In academic writing, expressions of enthusiasm, critique, or personal reflection can indicate human cognition \cite{kumarage2023stylometric,mahowald2024dissociating}. While AI models can mimic emotional language, their expressions typically lack intrinsic emotional depth and intentionality, often resulting in mechanical or contextually misplaced language.

\noindent  For instance, the following is a generated text from the dataset used in the \href{https://arxiv.org/html/2405.10129v1#:~:text=Table%201%3A,Across%20Six%20Categorie}{StyloAI} model, demonstrating a mechanical expression of emotion:
\begin{quote}
\texttt{"My car wont start, it doesnt turn over, it gives a clicking sound, keeps turning over but wont start, HELP."}
\end{quote}
The excerpt demonstrates the AI’s attempt to convey emotion through specific linguistic choices. The use of "HELP" adds urgency and signals distress. The short, fragmented phrases create a choppy rhythm that mirrors frustration, while the repetition of "won’t start" emphasises the speaker’s emotional intensity and sense of helplessness.

\paragraph{Psycholinguistic Interpretation:}
Human authors naturally embed emotional layers in their writing, drawing from lived experience, empathy, and reflective processes. These expressions evolve organically within the text, drawing on \textit{metacognition} and \textit{cognitive load}, showing contextual variation and dynamic sentiment progression.
An example of human-written text from the dataset used in the \href{https://arxiv.org/html/2405.10129v1#:~:text=Table%201%3A,Across%20Six%20Categorie}{StyloAI} model, that has a high emotion word count, but embeds it in layers:
\begin{quote}
\texttt{"If you find yourself struggling with patience as you work to change your reputation, remind yourself that your reputation isnt either mature or immature like a light switch is on or off, but is instead on a continuum that varies in maturity..."}
\end{quote}
In contrast, AI-generated sentiment is based on statistical patterns rather than experiential grounding. As a result, emotional expressions in AI texts often appear flat, stereotyped, or overly uniform.

\subsubsection{Readability Features}
Readability metrics assess the complexity and accessibility of a text, often combining indicators of lexical richness and syntactic structure. For example, \textit{Gunning fog}, which estimates the years of formal education needed to understand a text on the first reading, is one of the stylometric features extracted in this category. These features offer insight into a writer’s ability to adapt language for clarity and audience comprehension \cite{mindner2023classification,kumarage2023stylometric}. Human writers frequently adjust their writing based on reader expectations, while AI-generated texts may fluctuate between overly simple and unnecessarily complex content. The excerpt below of a generated text from the dataset used in the \href{https://arxiv.org/html/2405.10129v1#:~:text=Table%201%3A,Across%20Six%20Categorie}{StyloAI} model, that has a high value of Gunning fog, which means the text will be difficult to follow by a general audience:
\begin{quote}
\texttt{"The Cooperation Agreement PREAMBLE The Parties, Recalling their common will to promote, protect and ensure the rights and freedoms of persons with disabilities and to promote..."}
\end{quote}

\paragraph{Psycholinguistic Interpretation:}
Human authors regulate text complexity through deliberate planning, ongoing self-monitoring, and revision, balancing accessibility with depth. These processes reflect \textit{cognitive load}\textit{metacognitive awareness}, \textit{discourse-level planning}. In contrast, AI lacks true audience modelling capabilities and may produce writing that alternates unpredictably between overly dense and overly simplistic phrasing, without a clear rhetorical strategy.

\subsubsection{Named Entity Features}
Named entities such as people, organisations, locations, and dates offer important contextual clues about how authors ground their writing in real-world knowledge, personal experience, or domain-specific understanding \cite{mindner2023classification,kumarage2023stylometric}. Human writers often incorporate actual references, drawing from memory and context. In contrast, AI-generated texts may include fabricated, vague, or inconsistently integrated entities.

\paragraph{Psycholinguistic Interpretation:}
Human-authored writing reflects deliberate discourse planning, often integrating references to real events, individuals, or timelines. These references emerge from episodic memory, thematic coherence, and personal narrative strategies. These characteristics reference \textit{metacognition and discourse planning cognitive theories}. AI models, while capable of mimicking named entities, lack experiential grounding and may introduce irrelevant or inaccurate references due to limitations in factual recall or context awareness.

\subsubsection{Uniqueness Features}
Uniqueness refers to how distinct a piece of writing is from commonly observed linguistic patterns. This may involve novel word combinations (e.g., bigrams or trigrams), varied sentence structures, or stylistic originality \cite{kumarage2023stylometric}. Human authors typically produce more original phrasing, influenced by creativity, domain familiarity, and context-sensitive reasoning. AI-generated texts, on the other hand, often rely on high-probability sequences drawn from training data, resulting in more predictable or generic outputs.

\paragraph{Psycholinguistic Interpretation:}
Original expression in human writing is underpinned by semantic memory, flexible lexical retrieval, and the ability to adapt tone and structure to audience or purpose. These unique characteristics demonstrate \textit{lexical retrieval and discourse planning cognitive theories}. AI models, even when prompted for variety, generate phrases that are statistically probable but rhetorically shallow, lacking personal voice or adaptive strategy.

\section{Conclusion} \label{sec:conclusion}

This study demonstrates that integrating stylometric analysis with psycholinguistic theory offers an interpretable framework for distinguishing AI-generated from human-authored texts. By mapping 31 stylometric features to underlying cognitive processes, such as lexical retrieval, discourse planning, cognitive load management, and metacognitive self-monitoring, this research enhances the understanding of the psycholinguistic signatures embedded in human writing.
As AI tools become more integrated into educational and professional writing, it is essential to identify synthetic text while safeguarding the cognitive effort, originality, and ethical responsibility that characterise human authorship.

\bibliographystyle{splncs04}
\bibliography{AIED_Camera_Ready}

\appendix    

\begin{table}[!ht]
\centering
\scriptsize
\begin{tabular}{|l|p{3cm}|p{7cm}|p{3cm}|}
\hline
\rowcolor{blue!15}
\textbf{Feature} & \textbf{Description} & \textbf{Psycholinguistic Rationale} & \textbf{Stylometric Category} \\
\hline
\multirow{6}{*}{} WordCount & Total number of words in the text & WordCount reflects cognitive load balancing: more words often imply efficient idea generation and expression, as outlined by Sweller’s Cognitive Load Theory \cite{sweller1994cognitive}. & \multirow{6}{*}{Lexical Features} \\
\cline{1-3}
 UniqueWordCount & Number of distinct words used &  UniqueWordCount taps into lexical access and retrieval processes \cite{dell1992stages}. A broader range of vocabulary implies active retrieval from semantic memory and flexibility in language use, both requiring fluent cognitive control. &  \\
\cline{1-3}
 CharCount & Total number of characters &  CharCount correlates with discourse planning—longer texts usually present more cohesive arguments \cite{carrell1982cohesion}. &  \\
\cline{1-3}
AvgWordLength & Average length of words &  AvgWordLength indicates the level of vocabulary sophistication retrieved during writing \cite{dell1992stages}. Longer words may indicate access to more technical or abstract lexical items, but if cognitive load increases, writers might opt for simpler, shorter words to conserve cognitive resources \cite{de2010cognitive}. &  \\
\cline{1-3}
TTR & Ratio of unique words to total words &  A high TTR demonstrates dynamic lexical retrieval \cite{dell1992stages} and active self-monitoring \cite{flavell1979theories}. Low TTR can point to repetitive phrasing under mental strain. &  \\
\cline{1-3}
HapaxLegomenonRate & Proportion of words used only once & Hapax Legomena highlights creative, context-specific word use and metacognitive planning to avoid redundancy \cite{dell1992stages,flavell1979theories}. &  \\
\hline
\multirow{11}{*}{}SentenceCount & Total number of sentences & Captures how authors chunk ideas, balancing cognitive \cite{sweller1994cognitive} load and reader comprehension \cite{carrell1982cohesion}. & \multirow{11}{*}{Syntactic Features} \\
\cline{1-3}
AvgSentenceLength & Average number of words per sentence. & Longer sentences reflect higher planning effort; shorter ones can signal simplification under cognitive strain \cite{sweller1994cognitive}.  &  \\
\cline{1-3}
PunctuationCount & Total punctuation marks & PunctuationCount shows metacognitive control over sentence boundaries, rhythm, and readability \cite{flavell1979theories}. &  \\
\cline{1-3}
StopWordCount & Total count of common function words & StopWordCount balances grammatical fluency and cohesion \cite{carrell1982cohesion}. &  \\
\cline{1-3}
ComplexSentenceCount & Number of sentences with more than one clause & Complex sentences show layered discourse planning \cite{carrell1982cohesion} and advanced working memory management \cite{sweller1994cognitive}, allowing multiple ideas to be integrated cohesively. &  \\
\cline{1-3}
 QuestionCount & Total question marks & Questions reflect rhetorical engagement and audience-awareness strategies \cite{flavell1979theories}. &  \\
\cline{1-3}
ExclamationCount & Total exclamation marks & Exclamations signal emotional emphasis or spontaneity in discourse \cite{carrell1982cohesion}. &  \\
\cline{1-3}
ContractionCount & Number of contractions (e.g., don’t, can’t) & Contractions demonstrate stylistic awareness and adaptation to context, reflecting self-monitoring \cite{flavell1979theories} and flexible tone control. &  \\
\cline{1-3}
AbstractNounCount & Number of intangible concept nouns & Abstract nouns highlight conceptual abstraction abilities, engaging higher-level cognitive functions like thematic integration and abstract discourse planning \cite{carrell1982cohesion}. &  \\
\cline{1-3}
ComplexVerbCount & Less frequent verbs & The use of complex verbs reflects semantic sophistication and flexible lexical access \cite{dell1992stages}, often associated with subject-matter expertise. &  \\
\cline{1-3}
SophisticatedAdjectiveCount & Complex adjectives (e.g., “-ive,” “-ous”) & Highlights rich descriptive abilities tied to discourse planning and lexical retrieval, showing an intention to convey precision \cite{carrell1982cohesion}. &  \\
\hline
\multirow{4}{*}{}EmotionWordCount & Number of emotion-laden words & Emotion word usage reflects authentic emotional processing and discourse planning aimed at emotional engagement \cite{flavell1979theories}. & \multirow{4}{*}{Sentiment Features} \\
\cline{1-3}
 Polarity & Overall sentiment orientation & Polarity tracks shifts between positive and negative tones in argument flow \cite{flavell1979theories}. &  \\
\cline{1-3}
Subjectivity & Degree of opinion vs factual language & Subjectivity in text highlights personal stance and metacognitive involvement \cite{sweller1994cognitive}. &  \\
\cline{1-3}
VaderCompound & Overall sentiment score & Quantifies subtle emotional regulation throughout the text \cite{carrell1982cohesion}. &  \\
\hline
\multirow{2}{*}{}FleschReadingEase & Ease of reading score & Reflects cognitive effort in adjusting complexity for audience comprehension \cite{flavell1979theories} and working memory management \cite{sweller1994cognitive}. & \multirow{2}{*}{Readability Features} \\
\cline{1-3}
GunningFog & Education level needed to understand text & Indicates how writers balance depth of ideas against clarity and flow \cite{carrell1982cohesion}. &  \\
\hline
\multirow{4}{*}{}FirstPersonCount & First-person pronouns & Reflects self-referential awareness and narrative framing \cite{flavell1979theories}. & \multirow{4}{*}{Named Entity Features} \\
\cline{1-3}
DirectAddressCount & Direct reader engagement & Indicates audience-awareness and planning strategies \cite{carrell1982cohesion}. &  \\
\cline{1-3}
PersonEntities & Names of individuals & Taps into episodic memory retrieval \cite{dell1992stages,sweller1994cognitive} and authentic context integration. &  \\
\cline{1-3}
DateEntities & Temporal markers mentioned & Reflects the integration of personal or historical timelines in discourse \cite{dell1992stages,sweller1994cognitive}. &  \\
\hline
\multirow{2}{*}{}UniqueNgramCount & New word combinations (bigrams/trigrams) & Captures associative creativity and flexible lexical retrieval \cite{dell1992stages}, shaped by personal memory and discourse needs. & \multirow{2}{*}{Uniqueness Features} \\
\cline{1-3}
SyntacticVariety & Range of sentence structures & Reflects flexible discourse planning and adaptability \cite{carrell1982cohesion}. &  \\
\hline
\end{tabular}
\caption{Stylometric Features and Their Psycholinguistic Rationales, by Category}
\label{tab:merged-stylometric-features}
\end{table}

\end{document}